\documentclass{article}
\usepackage{spconf,amsmath,graphicx}
\usepackage{multirow}
\usepackage{amssymb,bm}
\usepackage{textcomp}
\usepackage{multirow}
\usepackage{tabularx}
\usepackage{mathtools}
\usepackage{array}
\usepackage{hyperref}
\usepackage{hhline}
\newcolumntype{P}[1]{>{\centering\arraybackslash}p{#1}}
\usepackage{diagbox}

\title{SCALABLE SENTIMENT FOR SEQUENCE-TO-SEQUENCE \\ CHATBOT RESPONSE WITH PERFORMANCE ANALYSIS}
\name{Chih-Wei Lee, Yau-Shian Wang, Tsung-Yuan Hsu, Kuan-Yu Chen,  Hung-Yi Lee, Lin-shan Lee}
\address{Graduate Institute of Communication Engineering, National Taiwan University \\
  {\footnotesize \tt \{k7922n, king6101, sivia89024, gary840212, tlkagkb93901106\}@gmail.com, lslee@gate.sinica.edu.tw}}
\begin{document}
\ninept
%
\maketitle
\begin{abstract}
Conventional seq2seq chatbot models only try to find the sentences with the highest probabilities conditioned on the input sequences, without considering the sentiment of the output sentences. Some research works trying to modify the sentiment of the output sequences were reported. In this paper, we propose five models to scale or adjust the sentiment of the chatbot response: persona-based model, reinforcement learning, plug and play model, sentiment transformation network and cycleGAN, all based on the conventional seq2seq model. We also develop two evaluation metrics to estimate if the responses are reasonable given the input. These metrics together with other two popularly used metrics were used to analyze the performance of the five proposed models on different aspects, and reinforcement learning and cycleGAN were shown to be very attractive. The evaluation metrics were also found to be well correlated with human evaluation.
\end{abstract}
\begin{keywords}
chatbot, dialogue, sequence-to-sequence, style-transfer, response generation
\end{keywords}
\vspace{-5mm}

\section{Introduction}
\label{sec:intro}
\vspace{-2mm}
Unlike goal-oriented dialogue systems \cite{lee2009example,wen2016network}, a chatbot is aimed to chat with human users on any subject domain of daily lives \cite{serban2016building,shang2015neural}.
The conventional chatbot is based on a seq2seq model \cite{vinyals2015neural} to generate meaningful responses given the user input. 
It is in general emotionless, and this is a major limitation of chatbots today because the emotion plays a critical role in human social interactions especially in chatting \cite{keltner1998emotion}. 
So we wish to train the chatbot to generate responses with scalable sentiment by setting the mode for chatting.
For example, for an input, ``How was your day today?'', the chatbot may respond, ``It is wonderful today'' or ``It is terrible today'' depending on the sentiment set, in addition to simply generating a reasonable response.
The mode can either be set by the developer or the user, or determined dynamically based on the context of the dialogue.
The techniques mentioned here may be extended to conversational style adjustment, so the machine may imitate the conversational style of someone the user is familiar with, to make the chatbot more friendly or more personal \cite{polzin2000emotion,hasegawa2013predicting}.

Substantial effort has been made focused on the conversation fluency and content quality of the generated responses, for example, by enriching the content diversity \cite{vijayakumar2016diverse,li2015diversity,li2016deep}, considering some additional information \cite{li2016persona}, addressing unknown words \cite{gu2016incorporating,eric2017copy} and so on. 
Some works tried to generate responses with controllable factors. 
The sentiment of a given sentence was successfully modified using non-parallel data \cite{shen2017style}. 
A chatbot which can change the style of responses by optimizing a given function related to the sentiment was also developed \cite{mueller2017sequence}. 
However, not too much work has been reported on scaling the sentiment of a chatbot, and how to properly evaluate a chatbot with adjustable sentiment is still a difficult problem \cite{shawar2007different,hung2009towards}. 

In this paper, we propose five approaches to scale the sentiment of chatbot responses and a set of evaluation metrics, and use these metrics to analyze the proposed approaches.
The five proposed approaches are: persona-based model, reinforcement learning, plug and play model, sentiment transformation network and cycleGAN, all based on the seq2seq model.
The set of four metrics to evaluate and analyze the different aspects of the chatbot responses are: two regarding if the responses are appropriate for the input; one regarding if the sentiment of the responses are properly modified; one regarding if the responses are grammatically good without considering the input. 
We then analyze the proposed approaches with these metrics, and find reinforcement learning and cycleGAN are very attractive.

\vspace{-3mm}

\section{Proposed Approaches} \label{sec:approach}

\vspace{-2mm}

Section \ref{subsec:seq2seq} briefly reviews the conventional seq2seq chatbot which was the basic model used by all the five proposed approaches presented in Section \ref{subsec:five}.
Below we assume we wish to make the chatbot response positive conditioned on the input, although it is easy to generalize the approaches to scalable sentiment.

\vspace{-3mm}

\subsection{Seq2seq Model (baseline)} \label{subsec:seq2seq}
\begin{figure}[h]
        \centering
        \includegraphics[width=\linewidth]{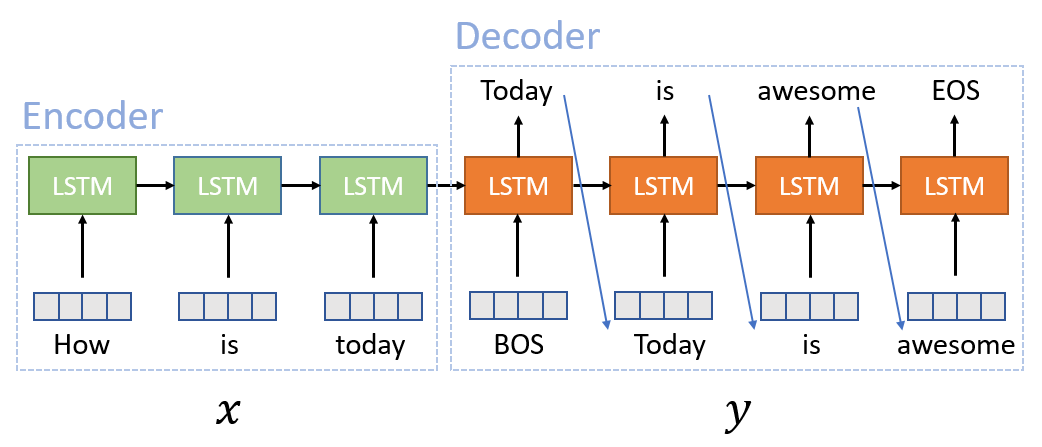}
        \caption{{\it Seq2seq model.}}
        \label{fig:seq2seq}
\end{figure}

Here we use attention-based seq2seq model \cite{luong2015effective} as in Figure \ref{fig:seq2seq} to train a simple chatbot using a corpus of dialogue pairs.
In all discussions here, $x$ is the input sentence to the seq2seq chatbot, and $y$ is the output of the seq2seq model.
$\hat{y}$ is the reference response in the training corpus.
In training phase, we input the sentence $x$ (a sequence of one-hot vectors) to the encoder, and the seq2seq model learns to maximize the probability of generating the sentence $\hat{y}$ given $x$. 

\vspace{-6mm}

\subsection{The Five Proposed Approaches} \label{subsec:five}

\subsubsection{Persona-Based Model} \label{subsec:persona}

\begin{figure}[h]
        \centering
        \includegraphics[width=\linewidth]{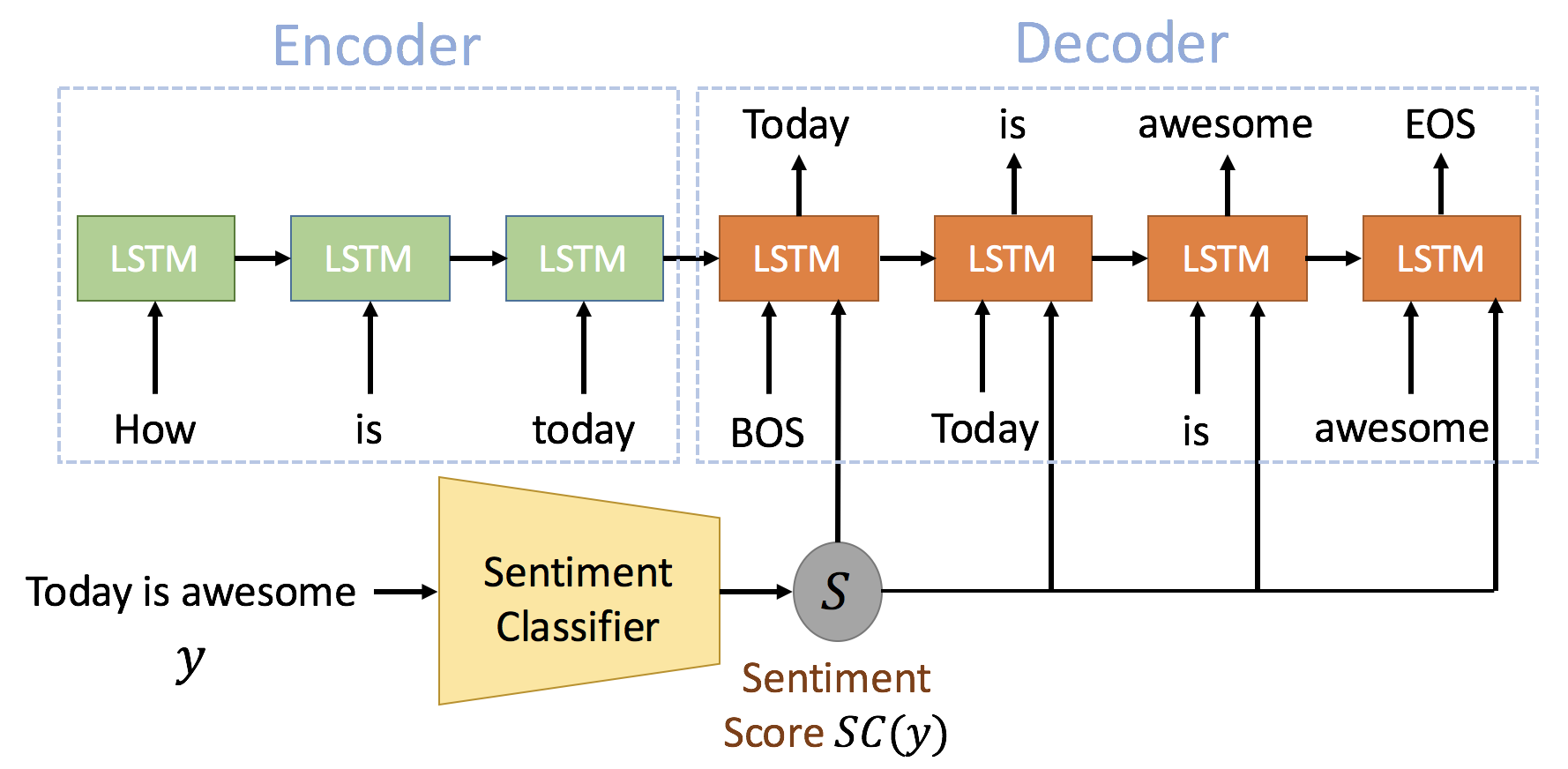}
        \caption{{\it Persona-based Seq2seq model}}
        \label{fig:persona}
\end{figure}

Persona-based model was originally proposed to generate sentences mimicking the responses of specific speakers \cite{li2016persona}.
It is very close to the seq2seq model, except adding extra information to the input of the decoder at each time step. 
In the original work \cite{li2016persona}, this extra information is the trained speaker embedding.
Here we replace the speaker embedding with a sentiment score (a scalar between $0$ and $1$) from a sentiment classifier as in Figure \ref{fig:persona}.
This sentiment classifier \cite{liu2012sentiment} is trained with a corpus of sentences with labeled sentiments to determine a sentence is positive or not.
The input of the classifier is a sentence $z$, and the output is a score $SC(z)$ between $0$ and $1$ indicating how positive the input is.
The input of the decoder at every time step is then the concatenation of the word embedding and a sentiment score.
During training the sentiment score of the reference sentence $SC(\hat{y})$ is used, and the decoder learns to generate the reference sentence.
For testing given the same input, we are able to scale the sentiment of the output by entering the desired sentiment score.

\vspace{-5mm}

\subsubsection{Reinforcement Learning} \label{subsec:reinforce}

Here exactly the same seq2seq chatbot as in Figure \ref{fig:seq2seq} is used, except we design a set of reward functions to scale the response sentiment with reinforcement learning.
Three components of the reward functions are developed as follow.

(1) \textit{Semantic Coherence 1}:
The response $y$ should be semantically coherent to the input $x$, in addition to being a good sentence.
So we pre-trained a different seq2seq model on a large dialogue corpus to estimate this semantic coherence with a probability $P_{coh}(y|x)$.
The first reward is therefore:
\begin{equation}\label{equ:lm}
R_1 = \frac{1}{N_y}\cdot log P_{coh}(y|x),
\end{equation}
where $x$ and $y$ denote the input and response of the baseline seq2seq chatbot (not the pre-trained seq2seq model), and $N_y$ is the length of $y$ for normalization.

(2) \textit{Semantic Coherence 2}: 
The semantic coherence mentioned above can be estimated in a completely different way.
We use the same dialogue corpus to train a RNN discriminator, in which two RNN encoders are used to represent the input $x$ and output $y$ as two embeddings, and these two embeddings are concatenated and followed by a fully connected layer to produce a score $D_{RNN}(x,y)$ between $0$ and $1$, to indicate if $x$ and $y$ are good dialogue pairs.
This score is therefore the second reward,
\vspace{-1mm}
\begin{equation}\label{equ:ch}
R_2 = D_{RNN}(x,y),
\end{equation}

(3) \textit{Sentiment Score}: 
The third reward is based on the sentiment classifier mentioned above in Section \ref{subsec:persona},
\vspace{-1mm}
\begin{equation}\label{equ:sc}
R_3 = SC(y),
\end{equation}
where $y$ is the seq2seq chatbot response.

The total reward is then the linear interpolation of the three rewards mentioned above,
\begin{equation}
R = \alpha \cdot R_1 + \beta \cdot R_2 + (1-\alpha-\beta) \cdot R_3
\end{equation}
where $\alpha$ and $\beta$ are hyper-parameters ranging from $0$ to $1$.
We employ the reinforcement learning algorithm with policy gradient \cite{sutton2000policy}.
\vspace{-3mm}
\subsubsection{Plug and Play Model} \label{subsec:plugandplay} 

\begin{figure}[h]
        \centering
        \includegraphics[width=\linewidth]{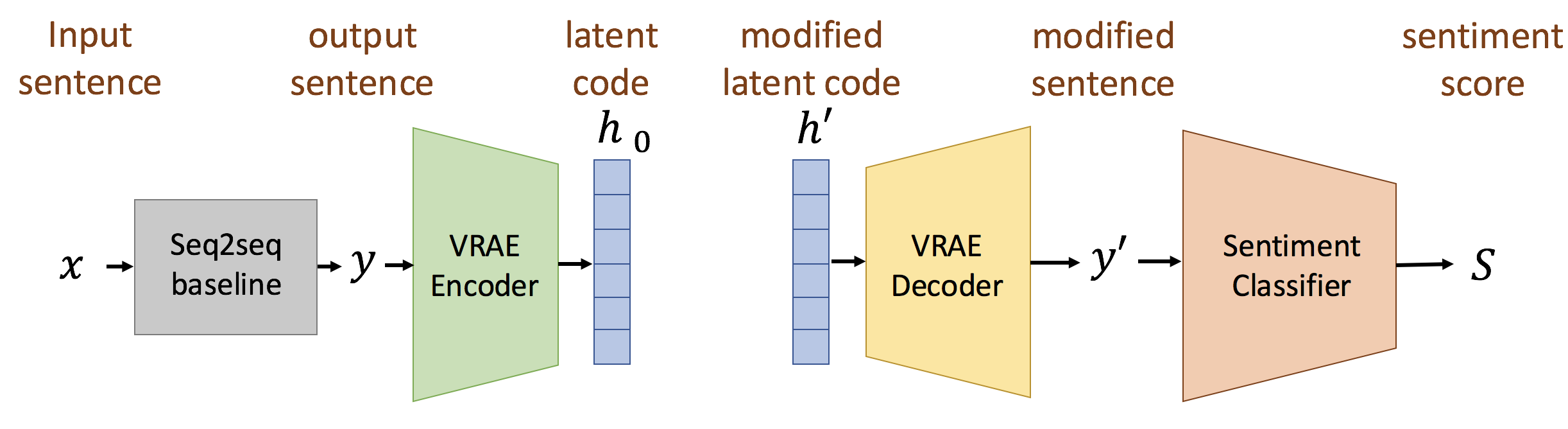}
        \caption{{\it Plug and play model. VRAE denotes variational recurrent auto-encoder}}
        \label{fig:plugandplay}
\end{figure}
\vspace{-3mm}
We borrow the concept of plug and play previously used to generate images \cite{nguyen2016plug} to generate dialogue response here,
as shown in Figure \ref{fig:plugandplay}. 
Here we pre-train a variational recurrent auto-encoder (VRAE) \cite{fabius2014variational} in addition using the same dialogue corpus. 
The encoder of VRAE on the left transforms a sentence $y$ into a fixed-length latent vector $h_0$, while the decoder of VRAE on the middle right generates a sentence $y'$ based on a vector $h'$. 
The encoder and decoder of VRAE is also jointly learned from the dialogue corpus for the chatbot.

\textit{The following steps happens on-line when the user enters a sentence.}
Given an input $x$, the seq2seq baseline first generates a response $y$, which is then encoded into a latent code $h_0$ by the VRAE encoder. 
Then the latent code $h_0$ is modified into $h^\prime$, based on the following equation:
\vspace{-2mm}
\begin{equation}
h' = argmax_{h} [\gamma\cdot SC( Decoder(h) ) - \delta\cdot MSE(h,h_0)],
\label{eq:pap}
\end{equation}
\hspace{-1mm}
where $SC$ denotes the sentiment classifier, $\gamma$ and $\delta$ are the weights of the loss function term and the regularization term.
The first term on the right-hand side of Eq.(\ref{eq:pap}), means we are looking for a code $h$ such that when it is decoded into a sentence $Decoder(h)$ using VRAE decoder, whose sentiment score $SC(Decoder(h))$ should be maximized.
The second term of Eq.(\ref{eq:pap}) prevents the code $h'$ being drifted too far from $h_0$.
To solve Eq.(\ref{eq:pap}), we calculate the gradient of the sentiment score with respect to the latent code $h$ and apply gradient ascent to the latent code iteratively
\footnote{Since the argmax layer between the decoder and sentiment classifier in $SC(Decoder(h))$ is non-differentiable, we use soft argmax \cite{kusner2016gans} to approximate argmax and then the gradient can be back-propagated throughout the whole network, from the sentiment classifier to the decoder.}, until the sentiment score output reaches a pre-defined value.
Because Eq.(\ref{eq:pap}) has to be solved on-line after the user enters an input sentence, this approach is more time consuming.

\vspace{-2mm}
\subsubsection{Sentiment Transformation Network} \label{subsec:transfernetwork}

This is very similar to the plug and play model previously mentioned in Section \ref{subsec:plugandplay} and Figure \ref{fig:plugandplay}, except here a sentiment transformation network $T_{\theta}$ with parameter set $\theta$ is learned, and $h'=T_{\theta}(h_0)$, or $T_{\theta}$ maps the latent code $h_0$ to a vector $h^\prime$, or to maximize the objective function with respect to $\theta$ instead of $h$.
So Eq.(\ref{eq:pap}) is replaced by:

\vspace{-5mm}
\begin{equation}
\theta^\prime = argmax_{\theta} [\epsilon \cdot SC(Decoder(T_{\theta}(h_0)))-\delta \cdot MSE(h,h_0)], 
\label{eq:stn}
\end{equation}
\hspace{-1mm}
where $\epsilon$ and $\delta$ are the weights of the loss function term and regularization term.
During training, we fix the weights of pre-trained VRAE and sentiment classifier but randomly initialize and then update the sentiment transformation network.
During testing, the code $h_0$ is adjusted by the sentiment transformation network $T_{\theta^\prime}$ learned in Eq.(\ref{eq:stn}), which generates the response.
\vspace{-4mm}
\subsubsection{CycleGAN (Cycle Generative Adversarial Network)} \label{subsec:cyclegan}
\vspace{-4mm}
\begin{figure}[h]
        \centering
        \includegraphics[width=\linewidth]{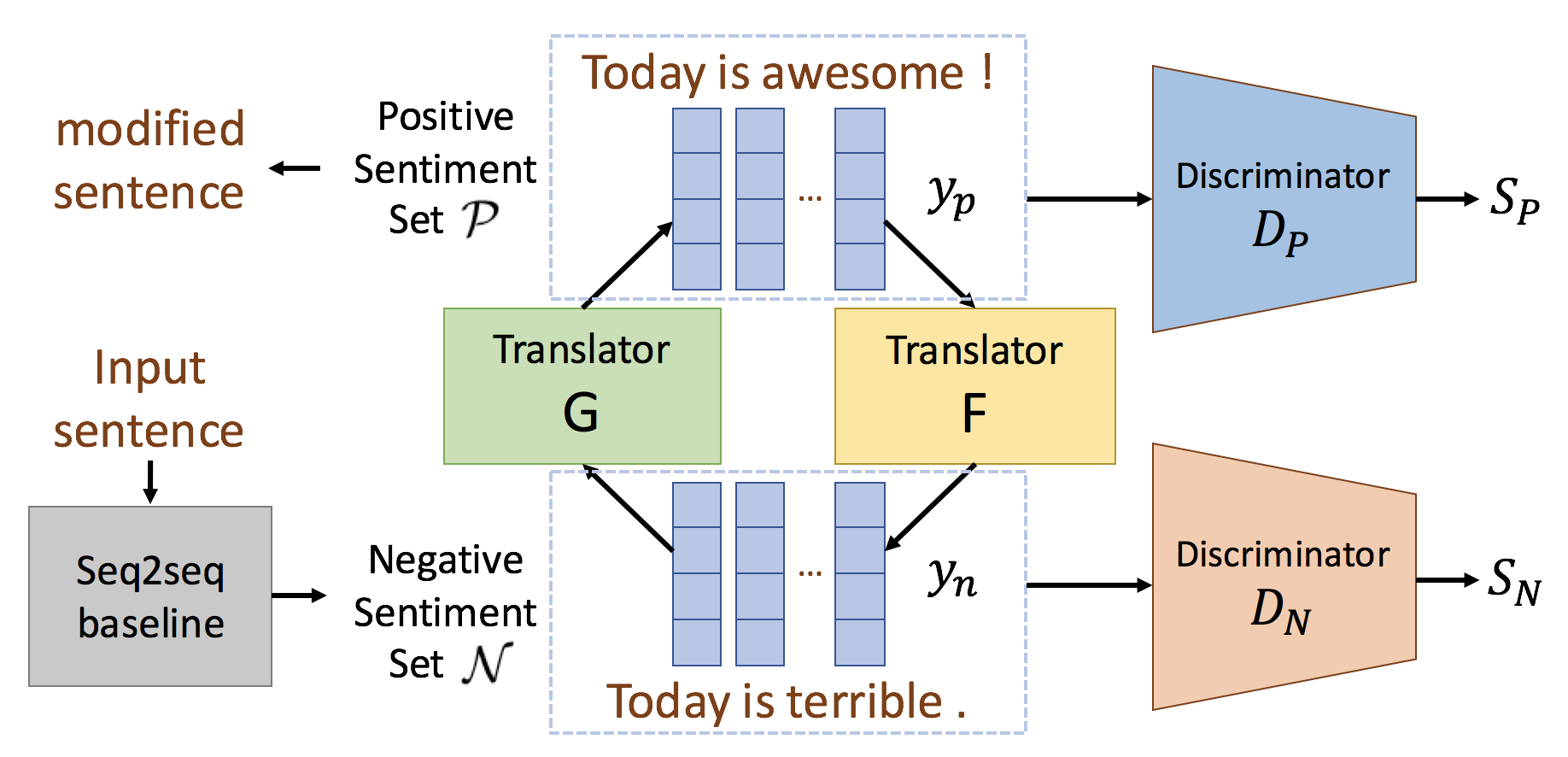}
        \caption{{\it CycleGAN Model for sentiment transformation. $F$ and $G$ are two translators respectively from positive to negative and negative to positive, and $D_P$ and $D_N$ are two discriminators respectively for positive and negative sentiment.}}
        \label{fig:cycleGAN}
\end{figure}
\vspace{-3mm}
Here we adopt the very powerful cycleGAN, which was shown very successful in image style transformation even without paired data \cite{zhu2017unpaired}. 
Here we show the way to use cycleGAN to transform the sentiment of sentences from negative to positive as in Figure \ref{fig:cycleGAN}. 
The model is trained with two sets of sentences in a corpus with labeled sentiments: positive sentiment set $\mathcal{P}$ and negative sentiment set $\mathcal{N}$. 
The sentences in the two sets are unpaired, or for a given sentence in  $\mathcal{P}$, it is not known which sentence in $\mathcal{N}$ corresponds to it.
We train two seq2seq translators, $G$ transforming a negative sentence $y_n$ into positive and $F$ from positive $y_p$ to negative.
We also train two discriminators, $D_{\mathcal{P}}$ and $D_{\mathcal{N}}$.
$D_{\mathcal{P}}$ and $D_{\mathcal{N}}$ takes a sequence of word embeddings as input and learn to distinguish whether this sequence is the word embeddings of a real sentence or generated by $G$ or $F$.
With the continuous word embeddings as the translator output, the gradient can be back-propagated from the discriminator to the translator. 
It's worth mentioning that $F$ and $G$ transform sequences of word embeddings to sequences of word embeddings.
We pre-train the word embedding model with Word2Vec \cite{mikolov2013efficient} and it is fixed during training the cycleGAN here. 
To transform the output sequence of word embeddings into a sentence, we simply select those words whose embedding has the highest cosine-similarity to each given word embedding in the sequence.

The concept of W-GAN \cite{arjovsky2017wasserstein} is used to train $D_{\mathcal{P}}$ and $D_{\mathcal{N}}$.
The loss function of the discriminator $D_{\mathcal{P}}$ is:
\vspace{-1mm}
\begin{equation}
L(D_{\mathcal{P}}) = D_{\mathcal{P}}(G(y_n)) - D_{\mathcal{P}}(y_p), \label{eq:dp}
\end{equation}

Where $y_n$ is a negative sentence sampled from $\mathcal{N}$, and $G(y_n)$ is the output of the translator $G$ taking $y_n$ as the input.
$D_{\mathcal{P}}$ learns to minimize Eq.(\ref{eq:dp}), or to give as low score to the translated output as possible (the first term on the right) and give as high score to real positive sentence $y_p$ as possible (the second term).
The loss function of the discriminators $D_{\mathcal{N}}$ is parallel to Eq.(\ref{eq:dp}),
\vspace{-1mm}
\begin{equation}
L(D_{\mathcal{N}}) = D_{\mathcal{N}}(F(y_p)) - D_{\mathcal{N}}(y_n). \label{eq:dn}
\end{equation}

As in Improved W-GAN, gradient penalty is applied here.
The loss functions for training the translators $G$ and $F$ are:
\vspace{-1mm}
\begin{equation}
\begin{aligned}
L(F) &= 2[MSE(y_p, G(F(y_p))) + MSE(y_n, F(G(y_n)))] \\
&\qquad - D_{\mathcal{N}}(F(y_p)), \label{eq:generatorG}
\end{aligned}
\end{equation}
\vspace{-1mm}
\begin{equation}
\begin{aligned}
L(G) &= 2[MSE(y_p, G(F(y_p))) + MSE(y_n, F(G(y_n)))] \\
&\qquad - D_{\mathcal{P}}(G(y_n)). \label{eq:generatorF}
\end{aligned}
\end{equation}

The first terms on the right-hand side of Eqs.(\ref{eq:generatorG}) and (\ref{eq:generatorF}) are the same.
Given a positive sentence $y_p$, after transformed into a negative sentence by $F$ and then transformed back to positive by $G$, it should be very close to the original sentence $y_p$.
Similar for the second terms.
The last terms of Eqs.(\ref{eq:generatorG}) and (\ref{eq:generatorF}) are different.
$F$ learns to generate output $F(y_p)$ considered by $D_{\mathcal{N}}$ as a real negative sentence, while $G$ learns to generate output $G(y_n)$ considered by $D_{\mathcal{P}}$ as a real positive sentence.
In this way the translators $F$, $G$ learn to transform the sentences from one sentiment (positive or negative) to the other.
Notice that the discriminators $D_\mathcal{P}$. $D_\mathcal{N}$ are jointly trained with the translators $F$, $G$.
During testing, for any chatbot output $y$, we simply use $G$ to transform it into a positive sentence $G(y)$.
\vspace{-5mm}
\section{Evaluation Metrics}
\vspace{-2mm}
Evaluation is always difficult in language generation, especially for chatbot.
Here we propose two metrics: sentiment coherence 1 and 2 (COH1, COH2) specially for chatbots, which give scores regarding whether the output sentence $y$ is a proper response to the input sentence $x$ or not.
They are in fact the Semantic Coherence 1 and 2 mentioned in Reinforcement Learning in Section \ref{subsec:reinforce} designed for the reward function.
But the seq2seq model and the RNN discriminator use to obtain these two scores were re-trained here therefore are slightly different models, although trained with the same corpus.

The third metric is the Sentiment Classifier Score (SCL) used to measure how positive the output sentence is.
This is in fact the sentiment classifier score $SC(y)$ used in the Persona-based model mentioned in Section \ref{subsec:persona}.
But the sentiment classifier used here is re-trained therefore is slightly different, although trained with the same corpus.
The fourth metric is the Language Model Score (LM) to check if the output sentence $y$ is a good sentence in terms of a language model \cite{mikolov2010recurrent}.
The language model used here was trained on the one billion word language modeling benchmark \cite{chelba2013one} using a two-layer GRU \cite{cho2014learning} model,

\begin{equation}
LM~Score = \frac{1}{N_y}\cdot log P(y),
\end{equation}
which is the language model probability $P(y)$ for a sentence $y$ but normalized with the sentence length $N_y$.
Note that the third and fourth metrics, SCL and LM, consider the output sentence $y$ only but not the input $x$.
The first and second metrics, COH1 and COH2, however, consider the output $y$ given the input $x$.
\vspace{-2mm}
\section{Experiments and Results}\label{sec:experiment}
\vspace{-3mm}
\subsection{Experimental Setup}

We trained all our models including the seq2seq baseline and the five proposed models using the Twitter chatting corpus available on Marsan-Ma's github repository \cite{Marsan-Ma} using tensorflow.
It contains about 3.7M of dialogue pairs. The whole corpus is split into training and validation set.
The latter included 28k dialogue pairs.
The sentiment classifier used in this work was trained from the twitter sentiment analysis corpus \cite{pak2010twitter}, which consists of 15M data with labeled sentiment ($0$ or $1$).
This corpus was also split into training and validation set.
The trained sentiment classifier reached $87\%$ of accuracy on validation set.
We trained six models, including the seq2seq baseline and the five models proposed, using the training set and evaluated these models using the validation set. 
The four evaluation metrics obtained are the average over the validation data. 
\vspace{-3mm}
\subsection{Experimental Results}

The results are listed in Table \ref{table:scores}. 
Notice that the seq2seq baseline in the first row was used in the five proposed models, therefore we didn't modify the sentiment for output of that model.
\vspace{-3mm}
\begin{table}[h]
\begin{tabular}{|P{2.8cm}|P{0.9cm}|P{0.9cm}|P{0.9cm}|P{0.9cm}|}
\hline
\multirow{2}{*}{\backslashbox{Model}{Metrics}} & \multicolumn{2}{c|}{Semantic Coh.} & 
Sent. & Lang. \\
\cline{2-5}
& COH1 & COH2 & SCL & LM\\
\hhline{|=|=|=|=|=|}
Seq2seq(baseline) &$\textnormal{-}0.755$&$0.762$&$0.543$&$\textnormal{-}1.465$\\
\hhline{|=|=|=|=|=|}
Persona-based&$\textnormal{-}1.961$&$0.710$&$\bf{0.870}$&$\textnormal{-}2.169$\\
\hline 
Reinforcement L.& $\bf{\textnormal{-}0.839}$& $\bf{0.792}$& $\bf{0.777}$ &$\bf{\textnormal{-}1.556}$\\
\hline
Plug and Play &$\textnormal{-}1.364$&$0.759$&$0.697$&$\textnormal{-}1.671$\\
\hline
Transformation Net&$\textnormal{-}1.566$&$0.743$&$0.624$&$\textnormal{-}1.996$\\
\hline
CycleGAN &$\bf{\textnormal{-}0.979}$&$\bf{0.764}$&$0.695$&$\bf{\textnormal{-}1.562}$\\
\hline
\end{tabular}
\caption{Evaluation results for the different models proposed, where COH1, COH2, SCL and LM stand for the four evaluation metrics: Semantic Coherence 1, Semantic Coherence 2, Sentiment Classifier Score and Language Model score respectively. The first row is for the seq2seq baseline.}
\label{table:scores}
\end{table}

\vspace{-6mm}

\subsection{Discussion on the Results}\label{subsec:discussion}

First consider the seq2seq baseline model.
The sentiment classifier score (SCL) is $0.543$ which is close to $0.5$.
This means the baseline model was more or less un-biased on positive or negative sentiments.
So it is a reasonable baseline.
Below we divide the discussions on the proposed models into two parts considering the different architectures of these models. 
\vspace{-3mm}
\subsubsection{Persona-Based Model and Reinforcement Learning}

These two models directly modified the seq2seq model's output, so the parameters of the seq2seq model were changed. 

For the persona-based model, the SCL score is extremely high but on the other hand its COH1 score is extremely low.
This is probably because we fed the model with the sentiment distribution from a pre-trained sentiment classifier, and as a result the model overfitted on this sentiment distribution. Therefore, it tries to output sentences not necessarily coherent to input, but with correct sentiment.
We noticed that its output very often contained two phrases in one sentence, hence the language model score is lower than the other models.

The Reinforcement Learning model performed better than all other models in three out of the four metrics: COH1, COH2 and LM, except for the SCL score.
This is because the reward $R_1$ and $R_2$ in Eqs.(\ref{equ:lm}), (\ref{equ:ch}) were in parallel with COH1 and COH2, and $R_1$ in Eq.(\ref{equ:lm}) also considered the word ordering which gave high LM score.
Its SCL score was also high (except not as high as the overfitted Persona-based model) because its reward $R_3$ is also in parallel with SCL, which made the output positive.
Due to the sampling mechanism, the reinforcement learning model was able to generate diverse responses which other models couldn't achieve.

From the data we also observed both the Persona-based and reinforcement learning models were able to make complicated changes to the output sentences which were rarely seen on other models.

\vspace{-3mm}
\subsubsection{Plug and Play, Sentiment Transformation Network and CycleGAN}

Instead of modifying the parameters of the seq2seq model, these three models modify the responses after they are generated by the seq2seq model.

Plug and Play and Sentiment Transformation Network both tried to modify the latent code of the sentences and they both used the gradient of the sentiment classifier. 
The sentiment classifier primarily considered the sentiment without really encoding the semantics of the sentences, hence when maximizing the sentiment classifier's output, the information from original input may be lost.
This is probably why COH1 and COH2 scores of these two models are both lower than most of the others. 

For CycleGAN, since the two translators directly output word embeddings carrying both sentiment and semantics, the translators were capable of finding the mapping between words like ``bad" to ``good", ``sorry" to ``thank", ``can't" to ``can". However, it could only change or delete some specific words but failed to make complex modification for the whole sentences. Since it only changes a few words of the original responses, the COH1, COH2 scores were not too far from the seq2seq baseline.

Some examples are shown on the following link: \url{goo.gl/X1PZLM}.
\vspace{-3mm}
\subsection{Human Evaluation}
\vspace{-3mm}
\begin{table}[h]
\begin{tabular}{|P{2.8cm}|P{1.35cm}|P{1.35cm}|P{1.35cm}|}
\hline
& Coherence & Sentiment & Grammar \\
\hhline{|=|=|=|=|}
Seq2seq(baseline) &$0.548$&$0.161$&$0.999$\\
\hhline{|=|=|=|=|}
Persona-based &$0.235$&$\bf{0.705}$&$0.746$\\
\hline 
Reinforcement L. & $\bf{0.346}$& $\bf{0.698}$& $\bf{0.925}$\\
\hline
Plug and Play &$0.150$&$0.483$&$0.430$\\
\hline
Transformation Net &$0.020$&$0.492$&$0.387$\\
\hline
CycleGAN &$\bf{0.435}$&$0.627$&$\bf{0.912}$\\
\hline
\end{tabular}
\caption{Human evaluation scores on the three questions regarding Coherence, Sentiment and Grammar. The average scores were normalized to from $0$ to $1$.}
\label{table:human}
\end{table}

\vspace{-4mm}

We performed subjective human evaluation with 30 subjects, all of whom were graduates students.
They were asked to answer three questions about the output sentences: (1) Coherence: Is the output sentence a good response to the input? (2) Sentiment: Is the output sentence positive? (3) Grammar: Is the output sentence grammatically correct?
They were asked to give scores ranging from $0$ to $5$, based on a few reference examples with given scores $1$, $3$, $5$ to scale the scores.
The average results (normalized to from $0$ to $1$) are listed in Table \ref{table:human}.

Since the subjective human evaluation questions are parallel to the objective machine evaluation scores, we calculate the Pearson correlation coefficients $\rho$ between Coherence, Sentiment and Grammar scores in Table \ref{table:human} with respect to COH1, SCL, and LM scores in Table \ref{table:scores}.
The results are $0.728$, $0.885$ and $0.543$ respectively.
This showed the machine evaluation metrics used here were well correlated to the human evaluation. 
\vspace{-3mm}
\section{Conclusion} \label{sec:conclusion}
\vspace{-2mm}
In this paper, we try to scale or adjust the sentiment of the chatbot response given the input. 
We propose five different models for this tasks, all based on the conventional seq2seq model.
We also propose two metrics to evaluate if the response is good for the given input.
After careful evaluation and analysis for the five proposed models on different aspects, we found among the five proposed models, Reinforcement Learning and CycleGAN were the most attractive.
The reinforcement learning was able to learn properly the different design goals and offer output sentences with good diversity.
The cycleGAN model primarily performed word mapping on the original response, so the output sentence quality was more or less preserved.
The Plug and Play model and Sentiment Transformation Network were not as successful at the moment, probably because it is not easy to modify the latent code of the sentences while preserving the semantics and sentence quality.

\newpage

\bibliographystyle{IEEEbib}
\bibliography{strings,refs}

\begin{thebibliography}{10}

\bibitem{lee2009example}
Cheongjae Lee, Sangkeun Jung, Seokhwan Kim, and Gary~Geunbae Lee,
\newblock ``Example-based dialog modeling for practical multi-domain dialog
  system,''
\newblock {\em Speech Communication}, vol. 51, no. 5, pp. 466--484, 2009.

\bibitem{wen2016network}
Tsung-Hsien Wen, David Vandyke, Nikola Mrksic, Milica Gasic, Lina~M
  Rojas-Barahona, Pei-Hao Su, Stefan Ultes, and Steve Young,
\newblock ``A network-based end-to-end trainable task-oriented dialogue
  system,''
\newblock {\em arXiv preprint arXiv:1604.04562}, 2016.

\bibitem{serban2016building}
Iulian~Vlad Serban, Alessandro Sordoni, Yoshua Bengio, Aaron~C Courville, and
  Joelle Pineau,
\newblock ``Building end-to-end dialogue systems using generative hierarchical
  neural network models.,''
\newblock in {\em AAAI}, 2016, pp. 3776--3784.

\bibitem{shang2015neural}
Lifeng Shang, Zhengdong Lu, and Hang Li,
\newblock ``Neural responding machine for short-text conversation,''
\newblock {\em arXiv preprint arXiv:1503.02364}, 2015.

\bibitem{vinyals2015neural}
Oriol Vinyals and Quoc Le,
\newblock ``A neural conversational model,''
\newblock {\em arXiv preprint arXiv:1506.05869}, 2015.

\bibitem{keltner1998emotion}
Dacher Keltner and Ann~M Kring,
\newblock ``Emotion, social function, and psychopathology.,''
\newblock {\em Review of General Psychology}, vol. 2, no. 3, pp. 320, 1998.

\bibitem{polzin2000emotion}
Thomas~S Polzin and Alexander Waibel,
\newblock ``Emotion-sensitive human-computer interfaces,''
\newblock in {\em ISCA tutorial and research workshop (ITRW) on speech and
  emotion}, 2000.

\bibitem{hasegawa2013predicting}
Takayuki Hasegawa, Nobuhiro Kaji, Naoki Yoshinaga, and Masashi Toyoda,
\newblock ``Predicting and eliciting addressee’s emotion in online
  dialogue,''
\newblock in {\em Proceedings of the 51st Annual Meeting of the Association for
  Computational Linguistics (Volume 1: Long Papers)}, 2013, vol.~1, pp.
  964--972.

\bibitem{vijayakumar2016diverse}
Ashwin~K Vijayakumar, Michael Cogswell, Ramprasath~R Selvaraju, Qing Sun,
  Stefan Lee, David Crandall, and Dhruv Batra,
\newblock ``Diverse beam search: Decoding diverse solutions from neural
  sequence models,''
\newblock {\em arXiv preprint arXiv:1610.02424}, 2016.

\bibitem{li2015diversity}
Jiwei Li, Michel Galley, Chris Brockett, Jianfeng Gao, and Bill Dolan,
\newblock ``A diversity-promoting objective function for neural conversation
  models,''
\newblock {\em arXiv preprint arXiv:1510.03055}, 2015.

\bibitem{li2016deep}
Jiwei Li, Will Monroe, Alan Ritter, Michel Galley, Jianfeng Gao, and Dan
  Jurafsky,
\newblock ``Deep reinforcement learning for dialogue generation,''
\newblock {\em arXiv preprint arXiv:1606.01541}, 2016.

\bibitem{li2016persona}
Jiwei Li, Michel Galley, Chris Brockett, Georgios~P Spithourakis, Jianfeng Gao,
  and Bill Dolan,
\newblock ``A persona-based neural conversation model,''
\newblock {\em arXiv preprint arXiv:1603.06155}, 2016.

\bibitem{gu2016incorporating}
Jiatao Gu, Zhengdong Lu, Hang Li, and Victor~OK Li,
\newblock ``Incorporating copying mechanism in sequence-to-sequence learning,''
\newblock {\em arXiv preprint arXiv:1603.06393}, 2016.

\bibitem{eric2017copy}
Mihail Eric and Christopher~D Manning,
\newblock ``A copy-augmented sequence-to-sequence architecture gives good
  performance on task-oriented dialogue,''
\newblock {\em arXiv preprint arXiv:1701.04024}, 2017.

\bibitem{shen2017style}
Tianxiao Shen, Tao Lei, Regina Barzilay, and Tommi Jaakkola,
\newblock ``Style transfer from non-parallel text by cross-alignment,''
\newblock {\em arXiv preprint arXiv:1705.09655}, 2017.

\bibitem{mueller2017sequence}
Jonas Mueller, David Gifford, and Tommi Jaakkola,
\newblock ``Sequence to better sequence: continuous revision of combinatorial
  structures,''
\newblock in {\em International Conference on Machine Learning}, 2017, pp.
  2536--2544.

\bibitem{shawar2007different}
Bayan~Abu Shawar and Eric Atwell,
\newblock ``Different measurements metrics to evaluate a chatbot system,''
\newblock in {\em Proceedings of the Workshop on Bridging the Gap: Academic and
  Industrial Research in Dialog Technologies}. Association for Computational
  Linguistics, 2007, pp. 89--96.

\bibitem{hung2009towards}
Victor Hung, Miguel Elvir, Avelino Gonzalez, and Ronald DeMara,
\newblock ``Towards a method for evaluating naturalness in conversational
  dialog systems,''
\newblock in {\em Systems, Man and Cybernetics, 2009. SMC 2009. IEEE
  International Conference on}. IEEE, 2009, pp. 1236--1241.

\bibitem{luong2015effective}
Minh-Thang Luong, Hieu Pham, and Christopher~D Manning,
\newblock ``Effective approaches to attention-based neural machine
  translation,''
\newblock {\em arXiv preprint arXiv:1508.04025}, 2015.

\bibitem{liu2012sentiment}
Bing Liu,
\newblock ``Sentiment analysis and opinion mining,''
\newblock {\em Synthesis lectures on human language technologies}, vol. 5, no.
  1, pp. 1--167, 2012.

\bibitem{sutton2000policy}
Richard~S Sutton, David~A McAllester, Satinder~P Singh, and Yishay Mansour,
\newblock ``Policy gradient methods for reinforcement learning with function
  approximation,''
\newblock in {\em Advances in neural information processing systems}, 2000, pp.
  1057--1063.

\bibitem{nguyen2016plug}
Anh Nguyen, Jason Yosinski, Yoshua Bengio, Alexey Dosovitskiy, and Jeff Clune,
\newblock ``Plug \& play generative networks: Conditional iterative generation
  of images in latent space,''
\newblock {\em arXiv preprint arXiv:1612.00005}, 2016.

\bibitem{fabius2014variational}
Otto Fabius and Joost~R van Amersfoort,
\newblock ``Variational recurrent auto-encoders,''
\newblock {\em arXiv preprint arXiv:1412.6581}, 2014.

\bibitem{kusner2016gans}
Matt~J Kusner and Jos{\'e}~Miguel Hern{\'a}ndez-Lobato,
\newblock ``Gans for sequences of discrete elements with the gumbel-softmax
  distribution,''
\newblock {\em arXiv preprint arXiv:1611.04051}, 2016.

\bibitem{zhu2017unpaired}
Jun-Yan Zhu, Taesung Park, Phillip Isola, and Alexei~A Efros,
\newblock ``Unpaired image-to-image translation using cycle-consistent
  adversarial networks,''
\newblock {\em arXiv preprint arXiv:1703.10593}, 2017.

\bibitem{mikolov2013efficient}
Tomas Mikolov, Kai Chen, Greg Corrado, and Jeffrey Dean,
\newblock ``Efficient estimation of word representations in vector space,''
\newblock {\em arXiv preprint arXiv:1301.3781}, 2013.

\bibitem{arjovsky2017wasserstein}
Martin Arjovsky, Soumith Chintala, and L{\'e}on Bottou,
\newblock ``Wasserstein gan,''
\newblock {\em arXiv preprint arXiv:1701.07875}, 2017.

\bibitem{mikolov2010recurrent}
Tomas Mikolov, Martin Karafi{\'a}t, Lukas Burget, Jan Cernock{\`y}, and Sanjeev
  Khudanpur,
\newblock ``Recurrent neural network based language model.,''
\newblock in {\em Interspeech}, 2010, vol.~2, p.~3.

\bibitem{chelba2013one}
Ciprian Chelba, Tomas Mikolov, Mike Schuster, Qi~Ge, Thorsten Brants, Phillipp
  Koehn, and Tony Robinson,
\newblock ``One billion word benchmark for measuring progress in statistical
  language modeling,''
\newblock {\em arXiv preprint arXiv:1312.3005}, 2013.

\bibitem{cho2014learning}
Kyunghyun Cho, Bart Van~Merri{\"e}nboer, Caglar Gulcehre, Dzmitry Bahdanau,
  Fethi Bougares, Holger Schwenk, and Yoshua Bengio,
\newblock ``Learning phrase representations using rnn encoder-decoder for
  statistical machine translation,''
\newblock {\em arXiv preprint arXiv:1406.1078}, 2014.

\bibitem{Marsan-Ma}
``Chat corpus,'' \url{https://github.com/Marsan-Ma/chat_corpus}.

\bibitem{pak2010twitter}
Alexander Pak and Patrick Paroubek,
\newblock ``Twitter as a corpus for sentiment analysis and opinion mining.,''
\newblock in {\em LREc}, 2010, vol.~10.

\end{thebibliography}

\end{document}